\documentclass{article}

\usepackage{microtype}
\usepackage{graphicx}
\usepackage{subfigure}
\usepackage{booktabs} % for professional tables
\usepackage{amsmath}
\usepackage{amssymb}
\usepackage{multirow}
\usepackage{afterpage}%
\usepackage{cuted}
\usepackage{flushend}

\usepackage{hyperref}

\usepackage[accepted]{mlsys2022}

\begin{document}

\twocolumn[
% \mlsystitle{A High Performance Implementation for vertically federated Random Forest}
\mlsystitle{An Efficient and Robust System for Vertically Federated Random Forest}
% \mlsystitle{On Efficient and Robust Federated Platform with Optimized Random Forest for Efficient and Robust for Vertically} Federated with Efficient Vertical Robust Random Forest

% unless you've provided the [accepted] option to the mlsys2022


\mlsyssetsymbol{equal}{*}

\begin{mlsysauthorlist}
\mlsysauthor{Houpu Yao}{equal,jd}
\mlsysauthor{Jiazhou Wang}{equal,jd}
\mlsysauthor{Peng Dai}{jd}
\mlsysauthor{Liefeng Bo}{jd}
\mlsysauthor{Yanqing Chen}{jd}
\end{mlsysauthorlist}

\mlsysaffiliation{jd}{JD Finance America Corporation}

\mlsyscorrespondingauthor{Jiazhou Wang}{jiazhou.wang3@jd.com}
\mlsyscorrespondingauthor{Yanqing Chen}{yanqing.chen@jd.com}

\mlsyskeywords{Federated Learning, }

\vskip 0.3in

\begin{abstract}
As there is a growing interest in utilizing data across multiple resources to build better machine learning models, many vertically federated learning algorithms have been proposed to preserve the data privacy of the participating organizations.
However, the efficiency of existing vertically federated learning algorithms remains to be a big problem, especially when applied to large-scale real-world datasets. 
% For thorough hyper-parameter tuning, rapid model iteration, and real-time inference
In this paper, we present a fast, accurate, scalable and yet robust system for vertically federated random forest. With extensive optimization, we achieved $5\times$ and  $83\times$ speed up over the SOTA SecureBoost model \cite{cheng2019secureboost} for training and serving tasks.
Moreover, the proposed system can achieve similar accuracy but with favorable scalability and partition tolerance. 
Our code has been made public to facilitate the development of the community and the protection of user data privacy.
% To facilitate the development of the community and the protection of user data privacy, our code has been made public at:
% \url{https://github.com/jzForPaper/rf_mlsys2022/tree/main/demos/random_forest}
\end{abstract}
]



\printAffiliationsAndNotice{\mlsysEqualContribution} % otherwise use the standard text.

\section{Introduction}
\label{sec:intro}

Machine learning techniques have achieved much success over the past decade. However, serious privacy concerns have also been raised over user data privacy while training these models \cite{yang2019federated,kairouz2019advances}. Such concerns are ever-increasing, especially over privacy critical applications such as biomedical engineering \cite{sheller2020federatedmedical}, self-driving cars \cite{pokhrel2020federatedselfdriving}, mobile phones \cite{mcmahan2017communication}, smart home devices \cite{dorri2017blockchainsmarthome}, and finance \cite{long2020federatedbanking}. Legislative authorities also started taking action to protect user privacy from potential data breaches. For example, in 2012, European Commission released the European Union’s General Data Protection Regulation (GDPR) for digital privacy protection.

% Generally speaking, there are four major categories of techniques that can be used to protect user privacy: Homomorphic Encryption (HE) \cite{cheon2017homomorphic}, Multi Party Computation (MPC) \cite{lindell2005MPC}, Differential Privacy (DP) \cite{friedman2010DP}, and Trusted Execution Environment (TEE) \cite{chen2020TEE}. 
% HE enables some specific computations to be done on encrypted data, such as addition or(and) multiplication. MPC involves designing secure multi-party computation protocols for original operations. 
% DP masks the original data by adding random noises, which reveals only dataset statistics. 
% TEE relies on specifically designed hardware to offer secure execution space, which is beyond our discussion.

To this end, federated learning is proposed as an emerging ideology that performs model training by utilizing data from multiple parties without threatening data privacy \cite{mcmahan2017communication}.
Depending on the configuration of data ownership, it can be categorized as vertically federated learning, horizontal federated learning, and federated transfer learning:
vertically federated learning corresponds to the case where parties have common users, but each party holds different user features. Horizontal federated learning corresponds to the case where participating parties have different user samples, but feature categories possessed by these parties are generally the same. Federated transfer learning corresponds to the case where few common user and feature categories are shared by participating parties.
By bridging multiple parties through federated learning, a much larger and diversified decentralized dataset can be utilized. With better dataset comes more powerful models, which can bring huge benefits to society.

\begin{figure}
    \centering
    \includegraphics[width=\columnwidth]{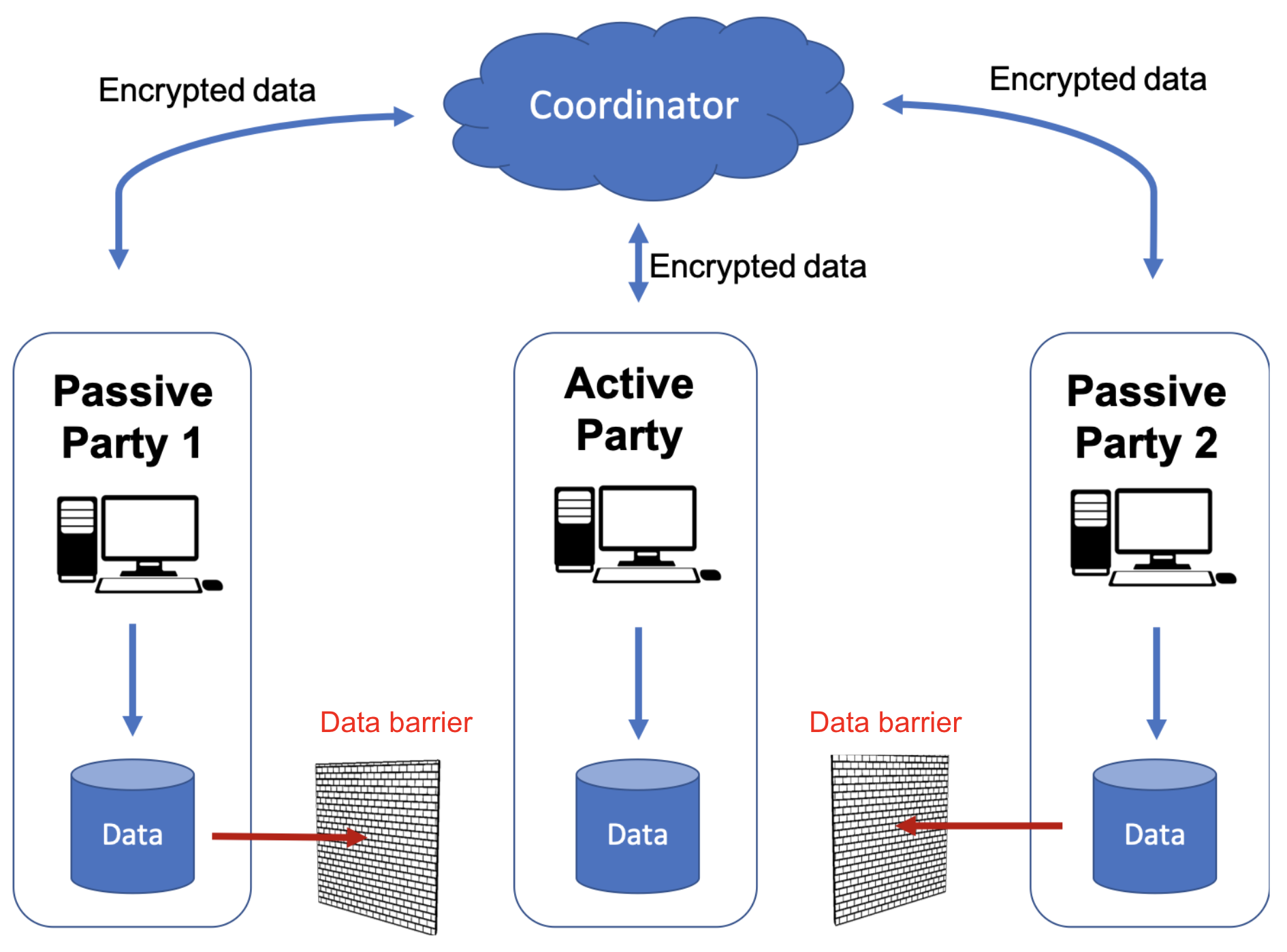}
    \caption{{Illustration of the vertically federated learning ideology, which breaks the data barrier between multiple organizations, so that jointly building a more powerful model could be made possible. The same set of users are served by these parties, and each party owns user features of different aspect.}}
    \label{fig:system_design}
\end{figure}

In this paper, we present a highly efficient and robust vertically federated random forest system. The system involves following aspects: (1) A novel federated random forest algorithm is designed for both efficient training and serving. (2) A novel distributed system is designed to utilize the parallelizablility of random forest, and to achieve high partition tolerance. (3) An efficient homomorphic cryptosystem is used to provide protection on data privacy. 
% Various optimization techniques have also been applied to further accelerate the speed.
% Furthermore, our model could achieve comparable accuracy and higher partition tolerance.

\begin{figure}
    \centering
    \includegraphics[width=\columnwidth]{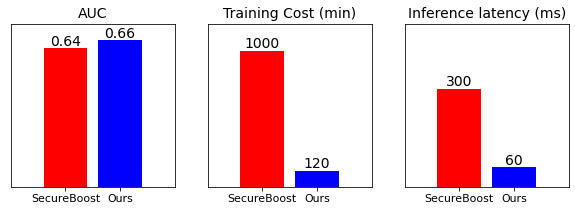}
    \caption{ The proposed method surpasses previous SOTA SecureBoost algorithm \cite{cheng2019secureboost}
    , proposed system .}
    \label{fig:performance_highlight}
\end{figure}

We highlight that, real-time inference is made possible for the first time through our proposed work. Compared with SOTA SecureBoost \cite{cheng2019secureboost}, the proposed system leads to has higher accuracy, less training time, and faster inference speed, as shown in Fig. \ref{fig:performance_highlight}.
The proposed federated learning system currently supports dataset size up to millions and has already provided hundreds of millions of calls in production without any system failure, proving its value in solving real-world problems.

% The rest of this paper is organized as follows: In the second section, we give a general definition to the problem we want to solve, briefly review existing techniques related to the proposed system. In the third section, we present our federated random forest designs, optimization techniques, and its robust and efficient service system. In the fourth section, several numerical experiments are conducted to demonstrate the performance of our system. We conclude the paper in the last section.

\section{Preliminaries}
\label{sec:related}

\subsection{Problem Definition}
\label{subsec:problem_def}

As a motivating example, we present a \emph{real world problem} that needs to be solved with federated learning ideology: To reduce the default risk in the credit authorization business, lenders (such as banks) must conduct comprehensive evaluations on the applicants. Furthermore, to make such evaluations more reliable, lenders would like to make use of various external data sources to build more powerful machine learning models. 

\emph{The challenge} to this problem comes from the following aspects: 
(1) How to utilize user data from other companies or organizations. Direct data copying can be problematic because accessing user data could be highly restricted due to privacy concerns and legislative considerations.
(2) How to make the model efficient during both training and inference stages. Faster training speed would be helpful for rapid model iteration, and faster inference speed provides a better customer experience.
(3) How to make the online service robust. It is highly desirable to have the system be partition tolerant when serving million-level users, otherwise it may bring a huge risk to the credit authorization business.

To this end, we design an efficient and robust federated learning system that allows multiple parties to jointly build machine learning models without leaking their private user data.
Without loss of generality, suppose there are $m$ participating parties, each possesses feature data  $X^{i} \in \mathbb{R}^{n_{i}\times d_{i}}$ where $n_{i}$ and $d_{i}$ are the number of samples and features on party $i$. 
We further assume there is an ``active party" that contains all the labels (lenders for example). The other participating parties that don't contain the label information are denoted as the ``passive party". 
\emph{Our goal} is to build an efficient and robust system to predict $y$ for the active party based on the features from all participating parties. In the meantime, the exact feature values are not revealed to the active party, and the label information owned by the active party is also kept secret. 

\subsection{Related Work}
% 

% (Vertically) FL related work
The federated learning ideology was first proposed by Google to protect the privacy of mobile phone users during keyboard inputting \cite{mcmahan2017communication}. 
Great success has been achieved by applying such ideology to its Gboard product: collaboratively learning a shared prediction model across devices, while keeping all user input history data locally. 
Since then, lots of work has been done to use federated learning to solve existing problems \cite{dayan2021naturecovid, dorri2017blockchainsmarthome,yang2020federatedrecommendation}.
As a special case, the vertically federated learning ideology is more suitable and often used in building cross-enterprise applications \cite{yang2019federated,liu2021fedlearn}, where different aspects of user features are owned by different enterprises.

% # Models
To perform different tasks, various machine learning models have been adapted into federated learning versions, such as linear model \cite{hardy2017private,yang2019federated}, decision tree \cite{vaidya2005privacy,vaidya2008privacy}, random forest \cite{liu2020federatedforest}, kernel methods \cite{gu2020federated}, XGBoost \cite{cheng2019secureboost}, and neural networks \cite{liu2020secure}. 
However, earlier model (such as the ones from \cite{vaidya2005privacy} or \cite{vaidya2008privacy}) usually suffers from security issues \cite{cheng2019secureboost}, or less efficient and scalable (such as the ones from \cite{liu2020federatedforest} or \cite{cheng2019secureboost}).

% System design
As federated learning has raised lots of attention over the recent years, researchers also start to investigate building efficient libraries \cite{he2020fedml, beutel2020flower} and systems \cite{bonawitz2019towards,bonawitz2019FLSys,hiessl2020industrial}. Several seminal work has also been done on designing better communication protocol \cite{konevcny2016federated}, optimization algorithms\cite{li2018federatedoptimization}, and improving model robustness \cite{konstantinidis2021byzshield}.
However, most of the existing research is targeting horizontal federated learning systems and vertical federated learning systems are still waiting to be tailored for industrial production.

% We have built a scalable production system for Federated pipeline. However, this training can easily be made to fail if some workers behave in an adversarial (Byzantine) fashion

% context as strong
% data similarity is assumed for all FL tasks \cite{hiessl2020industrial}

% has been done on horizontal federated learning.
% Learning in the domain of mobile
% devices, based on TensorFlow. \cite{bonawitz2019towards}

% {\color{red} How to design an efficient federated learning system has raised lots of interest over the past couple of years \cite{bonawitz2019FLSys}. To build efficient horizontal mobile system  \cite{bonawitz2019FLSys}, federated }

% \section{System Implementation}
% \label{sec:system}

% In this section, we first introduce how random forest can be trained under federated setting, i.e. label and feature are possessed by multiple parties, who will jointly build the model without leaking information to each other. In the second part, we introduce several optimization techniques to speed up the algorithm that enables real-time services. In the third part, we describe our distributed system that enables efficient training and robust serving of the proposed algorithm.

\section{Federated Random Forest}
\label{subsec:protocol}

Federated random forest is built upon classical random forest, whose final predictor is aggregated by multiple single tree predictors:
\begin{equation}
\begin{aligned}
F(x) =  \mathcal{A} \big( F_{1}(x_{1}), F_2(x_{2}), ..., F_{K}(x_{K}) \big)
% , \\ {\text \quad where \quad} x_{i}\subset x 
\end{aligned}
    \label{equation:RF_predictor}
\end{equation}
where $K$ is the total number of trees that will be trained, and $\{x_k \}^K_{k=1}$ are randomly sampled from the entire training set $X \in \mathbb{R}^{(n,d)}$ with $n$ samples and $d$ dimensional features. \(F(\cdot)\) is the random forest predictor, consists of a set of decision tree predictors \(\{ F_k \}^K_{k=1}\) each trained with a different sub dataset \(x_k\). The $\mathcal{A}$ operator is the corresponding aggregation strategy, such as mean, median etc.  Generally, the result of operator $\mathcal{A}$ will not change if a single (or even a couple of) $F_i(x_i)$ is missing.

What makes federated random forest different from classical random forest is that its training set $X$ is distributed on multiple parties and the information leakage between parties will not be allowed. 
As discussed in Sec.\ref{subsec:problem_def}, we denote these data partitions as $\{X^i\}^m_{i=1}$.
Nevertheless, similar to classical random forest, $\{F_k\}_{k=1}^K$ are still independent of each other and can be trained in parallel under federated learning settings. Such property will be explored in Sec.\ref{subsec:system} to build a robust service system.

In rest of this section, we introduce the training and real-time inference protocols for federated random forest.
The discussion on the protocol safety is included in Appendix \ref{appendix:protocol_safety}.

\subsection{Training protocol}

The training of federated random forest also involves the training of a collection of federated decision trees, and the training loss may vary for different tasks. Taking federated regression trees for example, choosing to minimize the variance gain at each tree node will lead to the following objective:
\begin{equation}
    L_{reg} = \frac{n_{L}E[Y_{L}]^{2} + n_{R}E[Y_{R}]^{2} - (n_{L}+n_{R})E[Y]^{2}}{n_{L} + n_{R}}
    \label{eq:regression_criteria}
\end{equation}
where $Y_L$ and $Y_R$ are the labels contained in the left and right sub-spaces created by the tree node, and $n_{L}$ and $n_{R}$ are the sub-space capacities.
Optimization on the training objective will be performed at each tree node level to find the best split threshold.Details on how Eq.\ref{eq:regression_criteria} is derived can be found in the Appendix \ref{appendix:desicion_tree_101}.

Similarly, we can optimize for the cross entropy loss for the classification task:
\begin{equation}
\label{eq:classification_criteria}
    L_{cls} = y log (p) - (1-y) log (1-p)
\end{equation}
where $p$ is the predicted probability generated by the model.

The overall training protocol of the proposed algorithm can be viewed as finite automata, whose control flow is given in Alg. \ref{alg:main}. The feature of instance space $I$ is distributed across all participating parties. The termination criteria $\mathcal{S}$ is used to perform early stopping and prevent over-fitting. It could include the maximum depth of the tree, the minimum number of samples required to split an existing node, and the minimum number of samples required to be at a leaf node.

\begin{algorithm}
   \caption{Main Control Flow (for active party)}
   \label{alg:main}
\begin{algorithmic}[1]
   \STATE {\bfseries Input:} Instance space $I$; encrypted label $ \left \langle Y \right \rangle$; termination criteria $\mathcal{S}$; number of trees $J$;
   \STATE {\bfseries Output:} Federated random forest $\{\mathcal{T}_j\}_{j=1}^{J}$
    \STATE {Perform ID alignment};
	\STATE \emph{Send encrypted label $ \left \langle Y \right \rangle $ to all parties};
   \WHILE{True}
		\STATE Call Alg. \ref{alg:sort} with encrypted label $ \left \langle Y \right \rangle $;
		\STATE Call Alg. \ref{alg:find} to \emph{retrieve encrypted data statistics $ \left \langle \mathbf{Y} \right \rangle$} and find the best split feature and threshold for current node;  
		\IF{$\mathcal{S}$ is satisfied}
		\STATE break;
		\ELSE
		\STATE Call Alg.  \ref{alg:split} to split the instance space into $I_L$ and $I_R$, create and \emph{broadcast the new node to all participating parties}; 
		\ENDIF
		\STATE Call Alg. \ref{alg:main} recursively to establish left subtree with $I_L$ and right subtree with $I_R$;
    \ENDWHILE
\end{algorithmic}
\end{algorithm}

In the initialization stage, the encrypted label will be broadcasted to all participating parties, and ID alignment will be performed.
{\color{black}Specifically, Diffie-Hellman algorithm \cite{li2010researchDH} will be used to perform private set intersection \cite{debnath2021securePSI} (PSI) and fulfill the ID alignment task.}
During the training process, Alg. \ref{alg:main} calls different subroutines located at each participating party. {\color{black} Such a remote calling process is done with gRPC under the proposed distributed system, which will be discussed in Sec.\ref{subsec:system}.
For better visualization, steps that involve data transfer are marked with the tilted font.}

The main purpose of Alg. \ref{alg:sort} is to obscure the raw data information by computing the statistics over a bin of data. 
The input includes hyper-parameter $P$ which specifies the number of bins to use for data obfuscation, local feature $X^i$ on the $i$-th party, and encrypted label $ \langle {Y} \rangle$. In the rest of this paper, we will use $\left \langle * \right \rangle $ to represent encrypted variables.
Such computation will be performed on all participating parties. It first loops over all features and sorts the encrypted label by each feature, then the aggregated feature mean $\mathbf{Y}$ will be computed. 
We set $P$ to be {around 100 } in later experiments. 
 
\begin{algorithm}
    \caption{Statistics Computation (for $i$-th party)}
    \label{alg:sort}
    \begin{algorithmic}[1]
        \STATE {\bfseries Input:} Homomorphically encrypted label $ \left \langle Y \right \rangle$; local feature $X^i$; number of bins $P$.
        \STATE {\bfseries Output:} Homomorphically encrypted statistics $ \langle \mathbf{Y} \rangle$.

        \FOR{$k = 0$ {\bfseries to} $d$}
            
            \STATE Sort $X_k^i$ (the $k$-th feature component of $X^i$);

            \STATE Compute the percentile: $V_{k} = \left \{v_{k,p}\right \}^{P}_{p=1}$, where $v_{k,p}$ are the $p$-th percentile value of $X_k^i$;

            \FOR{$p = 0$ {\bfseries to} $P$}
                \STATE  Get a bin of samples:
                
                $\mathcal{B} = \left \{ i|v_{k,p-1} \leq x_{j,k} < v_{k,p}, x_{j} \in X^i \right \}$
                \STATE Get the bin size: $n_{kp}=|\mathcal{B}|$;
                \STATE Compute homomorphically encrypted statistics: $\langle \mathbf{Y}_{kp} \rangle = \frac{1}{n_{kp}} \sum_{j \in \mathcal{B}} \left \langle {Y_j} \right \rangle$;
            \ENDFOR
            
        \ENDFOR
    \end{algorithmic}
\end{algorithm}

The best split feature and threshold will be selected in Alg. \ref{alg:find}. 
% It is selected from all participating parties' proposals, then be performed on the active party. 
The first step is to transfer all homomorphically encrypted data statistics $ \langle \mathbf{Y} \rangle$ obtained from Alg. \ref{alg:sort} to the active party.
Since only (encrypted) mean (or sum) for each bin will be sent from passive party to active party, the communication cost is actually not significant compared to the entire dataset.

\begin{algorithm}[h]
	\caption{Find Optimal Split (for active party)}
	\begin{algorithmic}[1]
		
		\STATE {\bfseries Input:} Instance space of the current node $I$, encrypted statistics $\left \{ \mathbf{Y}^{i}\right \}_{i=1}^{m}$, Minimum information gain $\epsilon$
		\STATE {\bfseries Output:} Optimal splitting feature $k_{opt}$ and the splitting threshold $v_{opt}$ 
		
		\FOR{$i = 0$ {\bfseries to} $m$}
		\STATE //\ enumerate all parties

    		\STATE \emph{Transfer $ \langle \mathbf{Y^i} \rangle$ from $i$-th party to the active party};
        
        		\FOR{$k = d_{0}$ {\bfseries to} $d_{i}$}
        		\STATE //\ enumerate all features
        
        		\FOR{$v = 0$ {\bfseries to} $v_{k}$}
            		\STATE //\ enumerate all threshold values
            		\STATE Decrypt $y^{i}_{kv} \gets  \mathbf{Y}^{i}_{kv}$;
            		\STATE Compute loss value $L_{kv}$ use Eq.\ref{eq:regression_criteria} or Eq.\ref{eq:classification_criteria};
            		\STATE Update $L_{opt}$, $k_{opt}$, $v_{opt}$ if $L_{kv}$ is lower; 
        		\ENDFOR
    		\ENDFOR
    		
		\ENDFOR
		
		\STATE  Current node will be split if $L_{opt} > \epsilon$;
		
	\end{algorithmic}
    \label{alg:find}
\end{algorithm}

Finally, Alg. \ref{alg:split} splits the current node based on the feature and threshold obtained. The instance space of the left and right child will be used to establish new nodes.
% instance space 
% and a new tree node is established.

\begin{algorithm}
	\caption{Update Instance Space (for all parties)}
	\begin{algorithmic}[1]
		\STATE {\bfseries Input:} Instance space of current node $I$; selected feature $k_{opt}$ and threshold $v_{opt}$;
		\STATE {\bfseries Output:} $I_L$ and $I_R$;
		\STATE Split $I$ into $I_L$ and $I_R$ according to $k_{opt}$ and $v_{opt}$
		\STATE Generate a record for the current branching, and \emph{broadcast to all participating parties};
	\end{algorithmic}
    \label{alg:split}
\end{algorithm}

Note that Alg. \ref{alg:sort} and Alg. \ref{alg:split} will be conducted on the active party when and only when the active party also has some feature information. In such cases, these algorithms will be performed locally using features in plaintext. 

\subsection{Real-time Inference Protocol}

As with random forest, federated random forest performs inference by comparing the feature value with the pre-trained threshold associated with each tree node. The comparison result is used to decide whether the left or right subtree will get activated until reaching the tree leaf.
What makes federated random forest different is that cross-party communication will be needed during the comparison. 
% How to speed up inference will be discussed in Sec.\ref{subsec:optimization}.

% The inference on federated random forest is similar to classical random forest in the sense that .
% What makes it different is that the feature can be located at different parties, and cross-party communication will be needed. 

% \subsubsection{O4: Optimized real time inference}

The inference with random forest model follows a top-down fashion: compare the feature value with the pre-trained threshold to determine which child node gets activated, which will be performed recursively till reaching the leaf node.
Under vertically federated learning settings, such a process becomes very expensive as cross-party communication emerges. Such communication delay can be a big problem when the tree depth increases.
For example, in our application, it is desirable to keep the service delay below 200 ms for a reliable customer experience.
However, there generally exists a 50ms delay in cross-party communication in our case, which constrains the tree height to be less than 4.
Due to the limit in maximum depth, the performance of random forest would also be limited.

\begin{algorithm}
	\caption{Optimized Federated Inference (for all parties)}
	\label{alg:real_time_inference}
	\begin{algorithmic}[1]
		\STATE Active party send inference initialization signal to passive parties.
		\STATE Passive parties compare their local features with the corresponding thresholds at each node, and send the result to the active party. 
		\STATE Active party ensemble all trees and perform inference.
	\end{algorithmic}
\end{algorithm}

We propose a new inference strategy to solve this problem.
Instead of performing the evaluation on the fly, the passive party will evaluate all possible paths in the decision trees it has, and send the information to the active party through a single communication. Based on the information received, the active party can ensemble the decision tree to perform inference locally. There are only 2 rounds of communications throughout the process, which corresponds to around 100ms delay. 
This strategy can be viewed as trading communication complexity with computational complexity, which is summarized in Alg. \ref{alg:real_time_inference}.
As a result, the inference could run in real-time without dropping any accuracy.

\begin{figure*}
 \centering
 \includegraphics[width=0.9\textwidth]{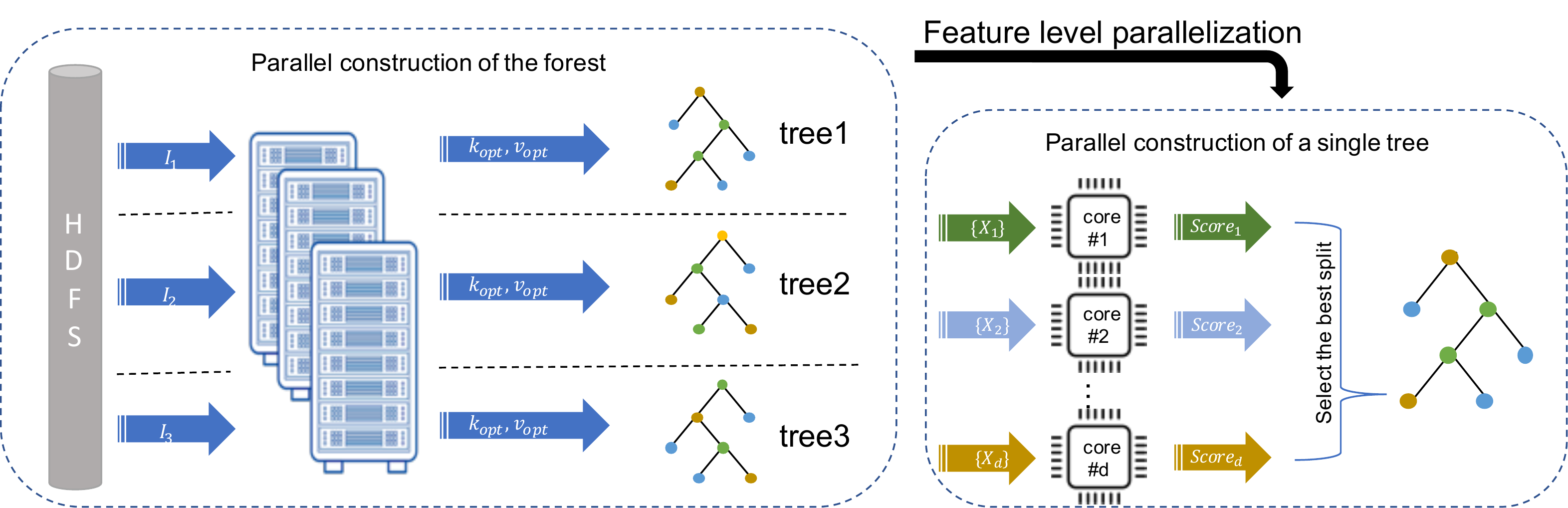}
 \caption{The distributed scheme of our system at tree (left) and feature (right) level. In our example, there are three trees built in parallel with three different computational units; and there are $d$ features for each tree, computed in parallel at each CPU core. Such design provide high partition tolerance to our system.}
 \label{fig:distributed_federated_randomforest}
\end{figure*}

\section{System Design}

The overall system performance comes from not only novel algorithm design but also from the high-performance system design. There are two major design considerations: (1) high partition tolerance, which ensures the service to be reliable, (2) low latency, which reduces customer waiting time. In this section, we will explore how such a system can be established.

\subsection{Distributed Service System}
\label{subsec:system}

% Aside from the asynchronous communication discussed in Sec. \ref{subsec:communication_optimization}, how to perform computation efficiently in a parallel or even distributed fashion is another direction to reduce system latency.
% When a single computational unit is down due to run-time exception or network failure, the overall system should not be impacted. 
% To this end, one distinguishing characteristic of random forest could be made use of: 

% the random forest brings is that the system performance will be impacted much when one tree (computational unit) is off-line

% When a single computational unit is down due to run-time exception or network failure, the overall system should not be impacted. 

% One major benefits that random forest brings is that the system performance will be impacted much when one tree (computational unit) is off-line. 

Similar to the classical random forest, federated random forest can easily be trained in parallel. There are two folds of parallelism: (1) all trees are independent of each other and can be built in parallel; (2) while splitting a node, the evaluation of each feature is independent of each other and can be done in parallel. As shown on the left part of Fig. \ref{fig:distributed_federated_randomforest}, each tree in the forest is built in parallel by different computational units. Furthermore, while building each tree, each feature can be evaluated in parallel using different CPU cores owned by each computational unit (shown in the right part of Fig. \ref{fig:distributed_federated_randomforest}).
Since the computation of building each tree and computing the statistics of each feature is completely independent of each other, {\color{black} the entire computation is fully decoupled and can be performed very well with MapReduce schema}.

% These nice properties could be fully explored to build an efficiently distributed service system at a large scale.

Furthermore, such parallelism offers great isolation between the computation of different trees, which means only one tree will be affected if a computation unit is down. Since the performance of the forest will not be influenced much by the absence of a single tree, high partition tolerance could be achieved.
Furthermore, since our system is initially designed for financial applications, high partition tolerance is required. With carefully designed system, off-line computational unit(s) should not impact the overall system performance.

% Taking the above-mentioned design requirements and analysis into consideration, we parallelize our distributed system at two different levels.

\subsubsection{Optimized communication}
\label{subsec:communication_optimization}

Heavy cross-party communication can be the major bottleneck in federated learning. To further improve the system efficiency, we optimize its communication process in this subsection.

Compared with plaintext, encryption significantly increases the message size, which leads to more time consumption in data transfer and eventually the overall training time. Thus, we send $ \langle {Y} \rangle$ in the initialization stage, which only happens once. During the training process, we only need to send encrypted statistics $ \langle \mathbf{Y} \rangle$, whose size is much smaller. 
% Therefore, we adopt several data compression techniques and asynchronization strategies to reduce this part of the cost.

When a new node is being created in the forest in Alg. \ref{alg:split}, $I_L$ and $I_R$ need to be broadcasted to all participating parties. As the sample space and the number of tree nodes can be very large, this step could potentially become another potential bottleneck to the training process. Here we discuss several ways to efficiently encode the instance space.
A naive way to serialize the index space is to direct jsonify the indices into a string, which we denote as the vanilla approach.
Alternatively, we can use an array of 1 and 0 bits to denote whether a sample has been selected or not, which we denote as bit encoding.
When the sampling rate is very small (close to leaf nodes), the bit string obtained from bit encoding can be super sparse. In this case, we can further encode the number of consecutive 0 bits instead based on the bit encoding result. We call this approach extreme bit encoding, which can be more efficient for an even smaller sampling rate.

% For example, 1024 bit RIAC ciphertext will encrypt a float to two large integers where each of them has more than 300 digits. 
% In practice, the total size of data transferred can be up to GB level. 

% {\color{red} ablation study result for async}
% won't wait for the response from a particular party after it sends out a message. Instead, it
% The coordinator will wait for response messages from all parties before starting the next computation step. 

Furthermore, we adopt an asynchronous communication strategy: instead of waiting for a response in Alg. \ref{alg:sort}, the system continues sending messages to sort different features. As illustrated in Fig. \ref{fig:optimized_communication}, with this strategy, Alg. \ref{alg:sort} can run in parallel on different parties and system idle time can be minimized.

\begin{figure}
    \centering
    \includegraphics[width=\columnwidth]{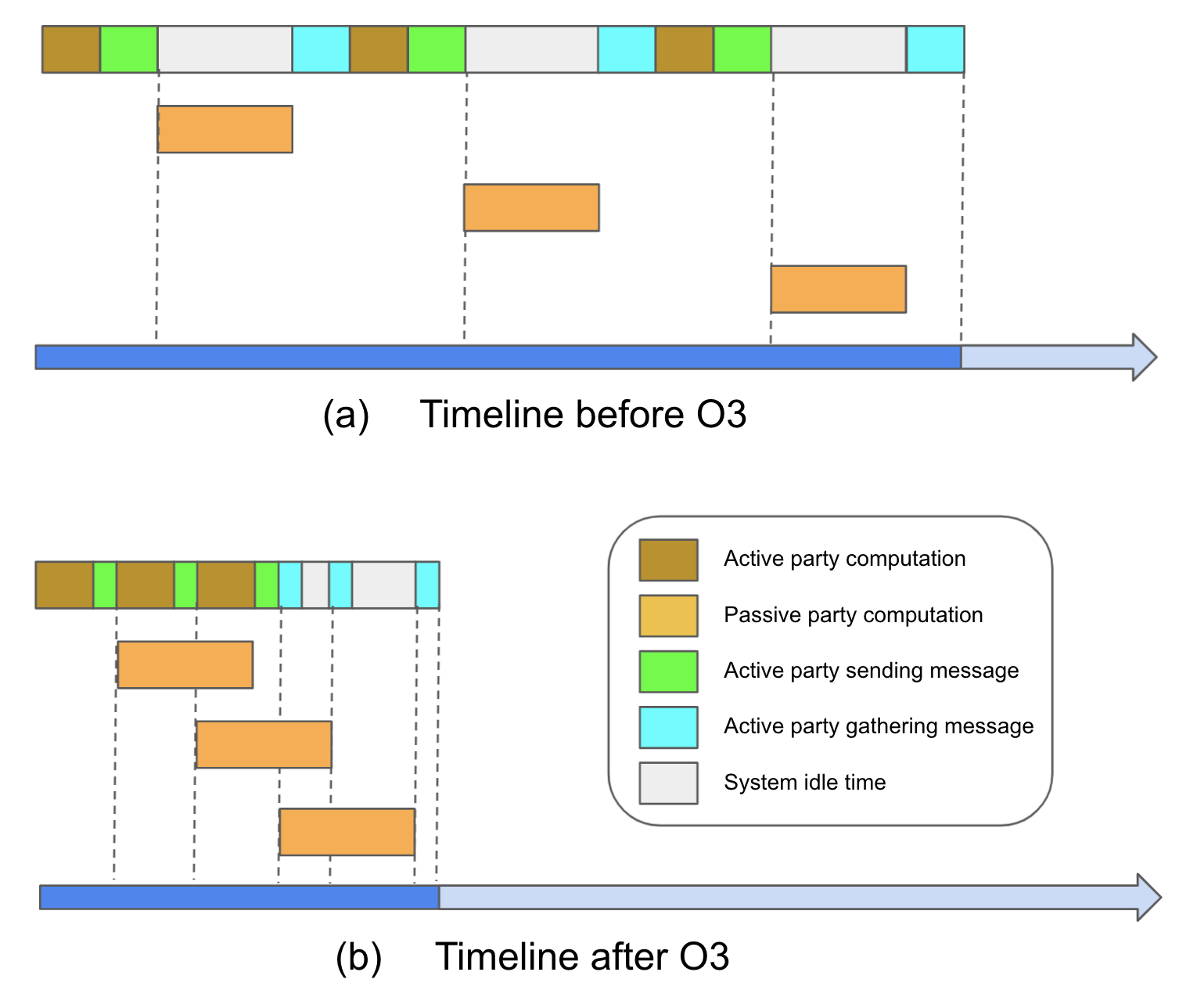}
    \caption{Illustration of proposed communication optimized. Asynchronous communication enables tasks to run in parallel, while better compression technique reduces the cost in sending and gathering messages across different participating parties. Both techniques contribute to the reduced overall time consumption.}
    \label{fig:optimized_communication}
\end{figure}

% as the base algorithm to build the federated learning platform, it brings another benefit: 
%  quality or 

% Another nice property of random forest 
% also be investigated.
% Furthermore, the system is expected to have ,

% partition tolerance is another major consideration factor in our system design. 

% Aside from better resource utilization, another benefit of such distributed design is that the resulting system could lead to a higher partition tolerance.

% And as stated in Alg. \ref{alg:find}, the best split for a node is obtained by selecting the best result over all features.

% Aside from the time consumption, we also need to consider the memory constrains in building the service platform. The system could raise Out of Memory (OOM) exception if the federated random forest becomes too complicated. To overcome the above mentioned issue and utilize the parallelizability of random forest, a high performance distributed platform is designed.

% \begin{figure}
%     \centering
%     \includegraphics[width=8cm]{./fig/***}
%     \caption{System design of proposed platform.}
%     \label{fig:system_design}
% \end{figure}

% each worker (core) only computes the mean statistics of one tree, or one feature of one tree, depending on the number of computing resources available.

\subsubsection{Optimized cryptosystem}
Homomorphic encryption is an ideal tool for designing federated learning algorithms. As a special form of encryption, it enables certain computations to be done on encrypted data without been decrypted first (such as RSA \cite{rivest1978RSA}, Paillier \cite{paillier1999public} or Gentry \cite{gentry2009FHE}). 

The fact that data lives in different parties (usually different geo locations) and the usage of homomorphic encryption in federated random forest brings extra time consumption in both communication and computation. Now we target at alleviating these issues to increase the system efficiency with modified cryptosystem and high performance computation library.

% Depending on the supported operator on ciphertext, popular homomorphic cryptosystems are generally categorized as semi-homomorphic and fully homomorphic. Semi-homomorphic encryption supports either multiplicative or additive operation on ciphertext (such as RSA \cite{rivest1978RSA} or  Paillier \cite{paillier1999public}), while fully homomorphic encryption (such as Gentry \cite{gentry2009FHE}) supports both. However, due to extremely high computational and data storage costs, current fully homomorphic encryption is still not of practical use in building real-world federated learning applications.

% \subsubsection{Homomorphic encryptosystem}

We adopt the Randomized Iterative Affine Cipher (RIAC)\footnote{{\color{black} The implementation is adapted from the open-source platform: \\ https://github.com/fedlearnAI/fedlearn-algo}} cryptosystem in our algorithm. The algorithm is included in the appendix for completeness.

\subsubsection{High performance multi-precision computation}

Computation with homomorphically encrypted ciphertext is generally composed of two operations: modular multiplication (mulmod) and modular exponentiation (powmod). These operations are applied to large integers (potentially several hundred digits).
To improve the computational efficiency of mulmod and powmod, we integrate the Multiple Precision Arithmetic (GMP) Library \cite{granlund1996gnu} into our system. GMP brings extra performance gain, which is analyzed in the Appendix.

\section{Numerical Results}
\label{sec:results}
In this section, we investigate the speed and accuracy of our federated random forest with several public datasets and datasets collected from our online business. Ablation studies are also conducted to help better understand the performance gains.

\subsection{Experimental Setup}

Five datasets were used in our experiments: a9a, gisette, and epsilon from LIBSVM \cite{libsvm}, 
and FK and HB gathered from our credit authorization business. 
The latter two datasets reflect real-world classification problems and include features such as user's credit history, purchase frequency, retailer distribution, etc. 
These datasets are binary labeled, representing user default or not.
A summary of these datasets are listed in Tab. \ref{table:benchmark_dataset_description}. 
All datasets are split into three parties: one active party with label and part of the feature, and two passive parties with the rest of the features randomly partitioned.

\begin{table}[h]
\caption{Dataset size ($n$) and feature dimension ($d$) of different datasets used in the experiment.}
\label{table:benchmark_dataset_description}
\vskip 0.15in
\begin{center}
\begin{small}
\begin{sc}
\begin{tabular}{lcccr}
\toprule
Datasets  & Feature dimension & Dataset size \\
\midrule
a9a  & 123 & 48,842 \\
gisette & 5,000 & 6,000 \\
epsilon & 2,000 & 400,000 \\
% \hline
HB & 115 & 77,000 \\
FK & 113 & 362,000 \\
\bottomrule
\end{tabular}
\end{sc}
\end{small}
\end{center}
\vskip -0.1in
\end{table}

The hardware we used to run the experiment has one Intel Xeon Gold 6130 CPU and 32G RAM.
Without further specification, each participating party is equipped with one such machine.
The network connection between participating parties has a speed of 300kB/s and a delay of 50 ms.
To fully exploit the parallelizability, the number of computation units in the distributed system is set to be the same as the number of trees in random forest.

\subsection{Results and Discussions}

\subsubsection{Model training}

We first compare the training time between the proposed algorithm and SOTA SecureBoost\cite{cheng2019secureboost}. As shown in Tab. \ref{table:results_time}, the proposed algorithm takes much less time on almost all datasets.
Moreover, for larger datasets like epsilon and FK, SecureBoost is not able to converge after the excessively long waiting time (15 hours and 5 hours respectively).
With much less training time, the proposed system enables more extensive hyper-parameter tuning and rapid model iteration, which could potentially lead to higher performance in accuracy.

\begin{table}[h]
\caption{Benchmark on training time (in minutes). We use ```$\textgreater X$'' to denote the training process does not converge after X minutes.}
\label{table:results_time}
\vskip 0.15in
\begin{center}
\begin{small}
\begin{sc}
\begin{tabular}{lccccr}
\toprule
DATA SET  & SECUREBOOST & OURS & BETTER? \\
\midrule
a9a  & 4 & 5 &  \\
gisette & 112 & 26 & $\surd$ \\
epsilon & $>$900 & 24 & $\surd$ \\
% \hline
HB & 117 & 21 & $\surd$ \\
FK & $>$300 & 60 & $\surd$ \\
\bottomrule
\end{tabular}
\end{sc}
\end{small}
\end{center}
\vskip -0.1in
\end{table}

% Several optimization techniques were proposed in Sec. \ref{subsec:optimization} to reduce the training time of federated random forest: GMP, RIAC, and optimized communication scheme.
% to demonstrate the benefits of these optimization techniques, wish which could bring some value to the community.

To fully understand the benefits of different strategies applied in our work, here we perform an ablation study with the FK dataset.
With all optimization techniques applied, the training time can be reduced from 35 hours to an hour. A breakdown of the performance improvement can be seen from Tab. \ref{table:improvement_roadmap}. 

\begin{table}[h]
\caption{Training time consumption on FK dataset with different levels of optimization applied. "+O1" means adding RIAC, "+O2" means further adding MPC, "+O3" means further adding optimized communication.}
\label{table:improvement_roadmap}
\vskip 0.15in
\begin{center}
\begin{small}
\begin{sc}
\begin{tabular}{lcccr}
\toprule
MODEL & TRAINING TIME & SPEEDUP\\
\midrule
Baseline & 2100 & 1$\times$ \\
+O1 & 420 & 5$\times$\\
+O2 & 210 & 10$\times$\\
+O3 & 60 & 35$\times$\\ 
\bottomrule
\end{tabular}
\end{sc}
\end{small}
\end{center}
\vskip -0.1in
\end{table}

In the next experiment, we compare the testing accuracy with AUC as the performance metric.
As shown in Tab. \ref{table:results_auc}, the proposed algorithm and SecureBoost lead to very similar performance.
The biggest difference is that, for SecureBoost, the AUC on the epsilon dataset is not obtained after 15 hours' training, thus labeled as ``N/A'' in the table. And SecureBoost also has a lower AUC on FK dataset as well, which is due to the fact that more training time would be needed  to fully converge.
Due to much faster training speed, the proposed algorithm is still in favor compared with SecureBoost as fast model iteration could be achieved. A more through investigation on the individual optimization techniques is included in the Appendix \ref{appendix:ablation_study}.

\begin{table}[h]
\caption{Performance comparison of different models. Benchmark on epsilon dataset. Numbers are measured in testing AUC. Unconverged models were denoted with ``N/A'', numbers adopted from \cite{lei2021stochastic} are denoted with $^*$, and ``--'' denotes the numbers couldn't be obtained due to privacy issues.}
\label{table:results_auc}
\vskip 0.15in
\begin{center}
\begin{small}
\begin{sc}
\begin{tabular}{lccccc}
\toprule
DATASET  & SECUREBOOST & OURS & Centralized \\
\midrule
a9a  & 0.89 & 0.89 &  $0.90^*$ \\
gisette & 0.99 & 0.99 & ${0.99}^*$ \\
epsilon & N/A & 0.79 &  0.79  \\
HB & 0.67 & 0.67 &   --  \\
FK & 0.64 & 0.66 &   -- \\
\bottomrule
\end{tabular}
\end{sc}
\end{small}
\end{center}
\vskip -0.1in
\end{table}

Furthermore, the impact that the proposed federated learning system has on model performance is also investigated. For the three public datasets, the AUC of models trained locally with the entire dataset is also listed in Tab. \ref{table:results_auc}. As shown in the ``centralized'' column, compared to the models trained locally, our federated random forest could achieve comparable performance while being able to protect data privacy.

\subsubsection{Model serving}

Once the federated random forest has been trained, it needs to be served efficiently and reliably with our system.
To investigate the inference speed, another experiment is performed with the proposed system, SecureBoost \cite{cheng2019secureboost} and unoptimized federated forest (similar to \cite{liu2020federatedforest}). 
For random forest, we set the forest to have 50 trees, and the maximum depth is set to be 10. {\color{black} As for SecureBoost, it has 25 trees and a depth of 6.}

As shown in Tab. \ref{table:inference_benchmark}, our system leads to a large improvement in inference speed over the existing work.
We observe that the optimization technique O4 reduces our system delay from 1000 ms to 120 ms, which makes real-time service possible. 
Besides, through batch processing of 100 queries per batch, our system composed of single physical machine for each party could achieved QPS up to 3000.
Furthermore, it is worth noting that this system can be scaled up easily for a larger random forest. Due to the fact that the trees are perfectly isolated, such tasks can be fulfilled by simply adding more machines. 

% , and there are 5 machines each runs at 30 TPS, which results to a total of 15k QPS. 

% Comparisons are made between

% The numbers for unoptimized federated forest were obtained from \cite{liu2020federatedforest}, {\color{red} where similar system were used}. 
% As shown in 

\begin{table}[h]
\caption{Benchmark on inference speed.}
\label{table:inference_benchmark}
\vskip 0.15in
\begin{center}
\begin{small}
\begin{sc}
\begin{tabular}{lcccr}
\toprule
MODEL & QPS & Latency ($\textit{s}$)\\
\midrule
SecureBoost & {\color{black} 0.1 } &  {\color{black} 10} \\
Unoptimized & {\color{black} $ 1$ } &  {\color{black} $ 1$} \\
Optimized (streaming) & {\color{black} 8} &  {\color{black} 0.12} \\
Optimized (Batching) & {\color{black} 3,000} &  {\color{black} 0.2} \\
\bottomrule
\end{tabular}
\end{sc}
\end{small}
\end{center}
\vskip -0.1in
\end{table}

% \begin{figure}
%     \centering
%     \includegraphics[width=5cm]{./fig/inference_speed_ablation.pdf}
%     \caption{Inference speed before and after proposed optimization.}
%     \label{fig:inference_speed_ablation}
% \end{figure}

\section{Conclusion}
An efficient and robust vertically federated random forest system is presented in this paper to preserve data privacy.  
For higher training and inference efficiency, several optimization techniques are utilized on the system design, cryptosystem, communication, and parallelization. Novel system design is also made to further improve the efficiency and robustness.
Ablation studies are provided to show the benefits of proposed optimization techniques, which we hope could help the community building even better practical federated learning systems in the future.
After extensive optimization, the proposed implementation for federated random forest algorithm could achieve favorable performance but much higher speed than SOTA benchmark on both open benchmark datasets as well as real-world business datasets.

\nocite{langley00}

\bibliography{mlsys}

\begin{thebibliography}{34}
\providecommand{\natexlab}[1]{#1}
\providecommand{\url}[1]{\texttt{#1}}
\expandafter\ifx\csname urlstyle\endcsname\relax
  \providecommand{\doi}[1]{doi: #1}\else
  \providecommand{\doi}{doi: \begingroup \urlstyle{rm}\Url}\fi

\bibitem[Beutel et~al.(2020)Beutel, Topal, Mathur, Qiu, Parcollet,
  de~Gusm{\~a}o, and Lane]{beutel2020flower}
Beutel, D.~J., Topal, T., Mathur, A., Qiu, X., Parcollet, T., de~Gusm{\~a}o,
  P.~P., and Lane, N.~D.
\newblock Flower: A friendly federated learning research framework.
\newblock \emph{arXiv preprint arXiv:2007.14390}, 2020.

\bibitem[Bonawitz et~al.(2019{\natexlab{a}})Bonawitz, Eichner, Grieskamp, Huba,
  Ingerman, Ivanov, Kiddon, Kone{\v{c}}n{\`y}, Mazzocchi, McMahan,
  et~al.]{bonawitz2019FLSys}
Bonawitz, K., Eichner, H., Grieskamp, W., Huba, D., Ingerman, A., Ivanov, V.,
  Kiddon, C., Kone{\v{c}}n{\`y}, J., Mazzocchi, S., McMahan, H.~B., et~al.
\newblock Towards federated learning at scale: System design.
\newblock \emph{arXiv preprint arXiv:1902.01046}, 2019{\natexlab{a}}.

\bibitem[Bonawitz et~al.(2019{\natexlab{b}})Bonawitz, Eichner, Grieskamp, Huba,
  Ingerman, Ivanov, Kiddon, Kone{\v{c}}n{\`y}, Mazzocchi, McMahan,
  et~al.]{bonawitz2019towards}
Bonawitz, K., Eichner, H., Grieskamp, W., Huba, D., Ingerman, A., Ivanov, V.,
  Kiddon, C., Kone{\v{c}}n{\`y}, J., Mazzocchi, S., McMahan, H.~B., et~al.
\newblock Towards federated learning at scale: System design.
\newblock \emph{arXiv preprint arXiv:1902.01046}, 2019{\natexlab{b}}.

\bibitem[Chang \& Lin(2011)Chang and Lin]{libsvm}
Chang, C.-C. and Lin, C.-J.
\newblock {LIBSVM}: A library for support vector machines.
\newblock \emph{ACM Transactions on Intelligent Systems and Technology},
  2:\penalty0 27:1--27:27, 2011.
\newblock Software available at \url{http://www.csie.ntu.edu.tw/~cjlin/libsvm}.

\bibitem[Cheng et~al.(2019)Cheng, Fan, Jin, Liu, Chen, and
  Yang]{cheng2019secureboost}
Cheng, K., Fan, T., Jin, Y., Liu, Y., Chen, T., and Yang, Q.
\newblock Secureboost: A lossless federated learning framework.
\newblock \emph{arXiv preprint arXiv:1901.08755}, 2019.

\bibitem[Dayan et~al.(2021)Dayan, Roth, Zhong, Harouni, Gentili, Abidin, Liu,
  Costa, Wood, Tsai, et~al.]{dayan2021naturecovid}
Dayan, I., Roth, H.~R., Zhong, A., Harouni, A., Gentili, A., Abidin, A.~Z.,
  Liu, A., Costa, A.~B., Wood, B.~J., Tsai, C.-S., et~al.
\newblock Federated learning for predicting clinical outcomes in patients with
  covid-19.
\newblock \emph{Nature medicine}, pp.\  1--9, 2021.

\bibitem[Debnath et~al.(2021)Debnath, Stǎnicǎ, Kundu, and
  Choudhury]{debnath2021securePSI}
Debnath, S.~K., Stǎnicǎ, P., Kundu, N., and Choudhury, T.
\newblock Secure and efficient multiparty private set intersection cardinality.
\newblock \emph{Advances in Mathematics of Communications}, 15\penalty0
  (2):\penalty0 365, 2021.

\bibitem[Dorri et~al.(2017)Dorri, Kanhere, Jurdak, and
  Gauravaram]{dorri2017blockchainsmarthome}
Dorri, A., Kanhere, S.~S., Jurdak, R., and Gauravaram, P.
\newblock Blockchain for iot security and privacy: The case study of a smart
  home.
\newblock In \emph{2017 IEEE international conference on pervasive computing
  and communications workshops (PerCom workshops)}, pp.\  618--623. IEEE, 2017.

\bibitem[Gentry et~al.(2009)]{gentry2009FHE}
Gentry, C. et~al.
\newblock \emph{A fully homomorphic encryption scheme}, volume~20.
\newblock Stanford university Stanford, 2009.

\bibitem[Granlund(1996)]{granlund1996gnu}
Granlund, T.
\newblock Gnu mp.
\newblock \emph{The GNU Multiple Precision Arithmetic Library}, 2\penalty0 (2),
  1996.

\bibitem[Gu et~al.(2020)Gu, Dang, Li, and Huang]{gu2020federated}
Gu, B., Dang, Z., Li, X., and Huang, H.
\newblock Federated doubly stochastic kernel learning for vertically
  partitioned data.
\newblock In \emph{Proceedings of the 26th ACM SIGKDD International Conference
  on Knowledge Discovery \& Data Mining}, pp.\  2483--2493, 2020.

\bibitem[Hardy et~al.(2017)Hardy, Henecka, Ivey-Law, Nock, Patrini, Smith, and
  Thorne]{hardy2017private}
Hardy, S., Henecka, W., Ivey-Law, H., Nock, R., Patrini, G., Smith, G., and
  Thorne, B.
\newblock Private federated learning on vertically partitioned data via entity
  resolution and additively homomorphic encryption.
\newblock \emph{arXiv preprint arXiv:1711.10677}, 2017.

\bibitem[He et~al.(2020)He, Li, So, Zeng, Zhang, Wang, Wang, Vepakomma, Singh,
  Qiu, et~al.]{he2020fedml}
He, C., Li, S., So, J., Zeng, X., Zhang, M., Wang, H., Wang, X., Vepakomma, P.,
  Singh, A., Qiu, H., et~al.
\newblock Fedml: A research library and benchmark for federated machine
  learning.
\newblock \emph{arXiv preprint arXiv:2007.13518}, 2020.

\bibitem[Hiessl et~al.(2020)Hiessl, Schall, Kemnitz, and
  Schulte]{hiessl2020industrial}
Hiessl, T., Schall, D., Kemnitz, J., and Schulte, S.
\newblock Industrial federated learning--requirements and system design.
\newblock In \emph{International Conference on Practical Applications of Agents
  and Multi-Agent Systems}, pp.\  42--53. Springer, 2020.

\bibitem[Kairouz et~al.(2019)Kairouz, McMahan, Avent, Bellet, Bennis, Bhagoji,
  Bonawitz, Charles, Cormode, Cummings, et~al.]{kairouz2019advances}
Kairouz, P., McMahan, H.~B., Avent, B., Bellet, A., Bennis, M., Bhagoji, A.~N.,
  Bonawitz, K., Charles, Z., Cormode, G., Cummings, R., et~al.
\newblock Advances and open problems in federated learning.
\newblock \emph{arXiv preprint arXiv:1912.04977}, 2019.

\bibitem[Kone{\v{c}}n{\`y} et~al.(2016)Kone{\v{c}}n{\`y}, McMahan, Yu,
  Richt{\'a}rik, Suresh, and Bacon]{konevcny2016federated}
Kone{\v{c}}n{\`y}, J., McMahan, H.~B., Yu, F.~X., Richt{\'a}rik, P., Suresh,
  A.~T., and Bacon, D.
\newblock Federated learning: Strategies for improving communication
  efficiency.
\newblock \emph{arXiv preprint arXiv:1610.05492}, 2016.

\bibitem[Konstantinidis \& Ramamoorthy(2021)Konstantinidis and
  Ramamoorthy]{konstantinidis2021byzshield}
Konstantinidis, K. and Ramamoorthy, A.
\newblock Byzshield: An efficient and robust system for distributed training.
\newblock \emph{Proceedings of Machine Learning and Systems}, 3, 2021.

\bibitem[Langley(2000)]{langley00}
Langley, P.
\newblock Crafting papers on machine learning.
\newblock In Langley, P. (ed.), \emph{Proceedings of the 17th International
  Conference on Machine Learning (ICML 2000)}, pp.\  1207--1216, Stanford, CA,
  2000. Morgan Kaufmann.

\bibitem[Lei \& Ying(2021)Lei and Ying]{lei2021stochastic}
Lei, Y. and Ying, Y.
\newblock Stochastic proximal auc maximization.
\newblock \emph{Journal of Machine Learning Research}, 22\penalty0
  (61):\penalty0 1--45, 2021.

\bibitem[Li(2010)]{li2010researchDH}
Li, N.
\newblock Research on diffie-hellman key exchange protocol.
\newblock In \emph{2010 2nd International Conference on Computer Engineering
  and Technology}, volume~4, pp.\  V4--634. IEEE, 2010.

\bibitem[Li et~al.(2018)Li, Sahu, Zaheer, Sanjabi, Talwalkar, and
  Smith]{li2018federatedoptimization}
Li, T., Sahu, A.~K., Zaheer, M., Sanjabi, M., Talwalkar, A., and Smith, V.
\newblock Federated optimization in heterogeneous networks.
\newblock \emph{arXiv preprint arXiv:1812.06127}, 2018.

\bibitem[Liu et~al.(2021)Liu, Tan, Wang, Zeng, Shan, Yao, Heng, Dai, Bo, and
  Chen]{liu2021fedlearn}
Liu, B., Tan, C., Wang, J., Zeng, T., Shan, H., Yao, H., Heng, H., Dai, P., Bo,
  L., and Chen, Y.
\newblock Fedlearn-algo: A flexible open-source privacy-preserving machine
  learning platform.
\newblock \emph{arXiv preprint arXiv:2107.04129}, 2021.

\bibitem[Liu et~al.(2020{\natexlab{a}})Liu, Kang, Xing, Chen, and
  Yang]{liu2020secure}
Liu, Y., Kang, Y., Xing, C., Chen, T., and Yang, Q.
\newblock A secure federated transfer learning framework.
\newblock \emph{IEEE Intelligent Systems}, 35\penalty0 (4):\penalty0 70--82,
  2020{\natexlab{a}}.

\bibitem[Liu et~al.(2020{\natexlab{b}})Liu, Liu, Liu, Liang, Meng, Zhang, and
  Zheng]{liu2020federatedforest}
Liu, Y., Liu, Y., Liu, Z., Liang, Y., Meng, C., Zhang, J., and Zheng, Y.
\newblock Federated forest.
\newblock \emph{IEEE Transactions on Big Data}, 2020{\natexlab{b}}.

\bibitem[Long et~al.(2020)Long, Tan, Jiang, and
  Zhang]{long2020federatedbanking}
Long, G., Tan, Y., Jiang, J., and Zhang, C.
\newblock Federated learning for open banking.
\newblock In \emph{Federated learning}, pp.\  240--254. Springer, 2020.

\bibitem[McMahan et~al.(2017)McMahan, Moore, Ramage, Hampson, and
  y~Arcas]{mcmahan2017communication}
McMahan, B., Moore, E., Ramage, D., Hampson, S., and y~Arcas, B.~A.
\newblock Communication-efficient learning of deep networks from decentralized
  data.
\newblock In \emph{Artificial Intelligence and Statistics}, pp.\  1273--1282.
  PMLR, 2017.

\bibitem[Paillier(1999)]{paillier1999public}
Paillier, P.
\newblock Public-key cryptosystems based on composite degree residuosity
  classes.
\newblock In \emph{International conference on the theory and applications of
  cryptographic techniques}, pp.\  223--238. Springer, 1999.

\bibitem[Pokhrel \& Choi(2020)Pokhrel and
  Choi]{pokhrel2020federatedselfdriving}
Pokhrel, S.~R. and Choi, J.
\newblock Federated learning with blockchain for autonomous vehicles: Analysis
  and design challenges.
\newblock \emph{IEEE Transactions on Communications}, 68\penalty0 (8):\penalty0
  4734--4746, 2020.

\bibitem[Rivest et~al.(1978)Rivest, Adleman, Dertouzos, et~al.]{rivest1978RSA}
Rivest, R.~L., Adleman, L., Dertouzos, M.~L., et~al.
\newblock On data banks and privacy homomorphisms.
\newblock \emph{Foundations of secure computation}, 4\penalty0 (11):\penalty0
  169--180, 1978.

\bibitem[Sheller et~al.(2020)Sheller, Edwards, Reina, Martin, Pati, Kotrotsou,
  Milchenko, Xu, Marcus, Colen, et~al.]{sheller2020federatedmedical}
Sheller, M.~J., Edwards, B., Reina, G.~A., Martin, J., Pati, S., Kotrotsou, A.,
  Milchenko, M., Xu, W., Marcus, D., Colen, R.~R., et~al.
\newblock Federated learning in medicine: facilitating multi-institutional
  collaborations without sharing patient data.
\newblock \emph{Scientific reports}, 10\penalty0 (1):\penalty0 1--12, 2020.

\bibitem[Vaidya \& Clifton(2005)Vaidya and Clifton]{vaidya2005privacy}
Vaidya, J. and Clifton, C.
\newblock Privacy-preserving decision trees over vertically partitioned data.
\newblock In \emph{IFIP Annual Conference on Data and Applications Security and
  Privacy}, pp.\  139--152. Springer, 2005.

\bibitem[Vaidya et~al.(2008)Vaidya, Clifton, Kantarcioglu, and
  Patterson]{vaidya2008privacy}
Vaidya, J., Clifton, C., Kantarcioglu, M., and Patterson, A.~S.
\newblock Privacy-preserving decision trees over vertically partitioned data.
\newblock \emph{ACM Transactions on Knowledge Discovery from Data (TKDD)},
  2\penalty0 (3):\penalty0 1--27, 2008.

\bibitem[Yang et~al.(2020)Yang, Tan, Zheng, Chen, and
  Yang]{yang2020federatedrecommendation}
Yang, L., Tan, B., Zheng, V.~W., Chen, K., and Yang, Q.
\newblock Federated recommendation systems.
\newblock In \emph{Federated Learning}, pp.\  225--239. Springer, 2020.

\bibitem[Yang et~al.(2019)Yang, Liu, Chen, and Tong]{yang2019federated}
Yang, Q., Liu, Y., Chen, T., and Tong, Y.
\newblock Federated machine learning: Concept and applications.
\newblock \emph{ACM Transactions on Intelligent Systems and Technology (TIST)},
  10\penalty0 (2):\penalty0 1--19, 2019.

\end{thebibliography}
\bibliographystyle{mlsys2022}

%%%%%%%%%%%%%%%%%%%%%%%%%%%%%%%%%%%%%%%%%%%%%%%%%%%%%%%%%%%%%%%%%%%%%%%%%%%%%%%
%%%%%%%%%%%%%%%%%%%%%%%%%%%%%%%%%%%%%%%%%%%%%%%%%%%%%%%%%%%%%%%%%%%%%%%%%%%%%%%
% SUPPLEMENTAL CONTENT AS APPENDIX AFTER REFERENCES
%%%%%%%%%%%%%%%%%%%%%%%%%%%%%%%%%%%%%%%%%%%%%%%%%%%%%%%%%%%%%%%%%%%%%%%%%%%%%%%
%%%%%%%%%%%%%%%%%%%%%%%%%%%%%%%%%%%%%%%%%%%%%%%%%%%%%%%%%%%%%%%%%%%%%%%%%%%%%%%
\newpage
\appendix

\section{Equations on federated trees}
\label{appendix:desicion_tree_101}

% \subsection{Classification}

Suppose $I$ is the instance space that contains data label $y$, $I_{L}$ and $I_{R}$ are the sub-spaces split by the tree node, $n_{L}$ and $n_{R}$ are the corresponding space capacities, $y_L$ and $y_R$ are the labels contained in the left and right sub-spaces. The regression loss for variance reduction can be formulated as:

\begin{equation}
    \begin{aligned}
    & Score 
    = Var[y] - \frac{|I_{L}|}{|I|}Var[y_{L}] - \frac{|I_{R}|}{|I|}Var[y_{R}] \\
     = &  E[Y^{2}] -E[Y]^{2} - \frac{n_{L}}{n_{L} + n_{R}}(E[Y_{L}^{2}] - E[Y_{L}]^{2}) \\ 
    &  - \frac{n_{R}}{n_{L} + n_{R}}(E[Y_{R}^{2}] - E[Y_{R}]^{2}) \\
    = & \frac{\sum Y^{2}}{n_{L} + n_{R}} - \frac{(\sum Y)^{2}}{(n_{L} + n_{R})^{2}} - \frac{\sum Y_{L}^{2}}{n_{L} + n_{R}}  \\
    & + \frac{(\sum Y_{L})^{2}}{n_{L}(n_{L} + n_{R})} - \frac{\sum Y_{R}^{2}}{n_{L} + n_{R}} + \frac{(\sum Y_{R})^{2}}{n_{R}(n_{L} + n_{R})} \\
    = & \frac{1}{N_{L} + N_{R}}[\frac{(\sum Y_{L})^{2}}{N_{L}} + \frac{(\sum Y_{R})^{2}}{N_{R}} - \frac{(\sum Y)^{2}}{N_{L} + N_{R}}] \\
     =&  \frac{1}{n_{L} + n_{R}}[n_{L}E[Y_{L}]^{2} + n_{R}E[Y_{R}]^{2} - (n_{L}+n_{R})E[Y]^{2}]
    \end{aligned}
    \label{eq:regression_criteria_ext}
\end{equation}

\section{Protocol Safety}
\label{appendix:protocol_safety}

% {Safety} is of primary importance when designing any federated learning algorithms.
% There are two potential threats to consider in practice: (1) will the raw data be safe under man-in-the-middle attach during the transfer process, and (2) 
% {\color{red} We will be focused on the second type of threat}, and consider the system safety under
In this part, we will analyze if the proposed federated random forest is secure.
% , i.e., if one party be able to derive parties' data based on the information it received.
The honest-but-curious (HBC) adversary will be considered: the participant does not deviate from the protocol but will attempt to learn all possible information from the received messages.
The system safety will be analyzed from two aspects: (1) Label security: will the passive party be able to derive the label information owned by the active party, and (2) Feature security: will the active party be able to derive feature information owned by the passive party.

% We highlight that the proposed algorithm suits all the safety analysis in \cite{cheng2019secureboost}.

% in a modified version: First active party creates a "local tree" with its features, then creates a regression forest with the passive parties using the residual from the local tree as the new labels.

% We refer to \cite{cheng2019secureboost} for the safety analysis, where the safety of federated tree-based models are investigated.

% Techniques to improve safety will also be discussed.

There are four data transfers during the training process: (1) In Alg. \ref{alg:main}, the active party will send the encrypted label to passive party upon training starting; (2) In Alg. \ref{alg:find}, passive parties will send $ \langle \mathbf{Y} \rangle$ ( accumulated encrypted label value by percentiles) to the active party; { (3) In Alg. \ref{alg:split}, active party will send the best split feature, threshold, and the samples space of left and right subtree to all participating parties.}

\textbf{{Label security}:}
Since labels are encrypted before sending out by the active party, brutal force decryption can be difficult or infeasible.
% Furthermore, the leaf node information are only stored at the active party, thus the prediction won't be obtained by the passive party either.
%  that leaf nodes' parent could probe the leaf nodes' information
However, a possible threat was pictured in \cite{cheng2019secureboost}: Consider a node owned by one passive party whose children nodes are tree leaves owned by the active party.
Since the sample space of the leaf nodes is decided by the parent node (in the passive party), then the passive party knows the sample space of these two leaf nodes. Under such situations, the passive party can easily guess the prediction result of these two leaf nodes, and the guessing accuracy is based on the purity of the leaf nodes. 
In the meantime, since random forest tends to over-fit on each individual tree, the risk that the prediction of a single tree gets detected is also higher. 
We adopted the {\color{black}"Completely SecureBoost"} \cite{cheng2019secureboost} ideology to alleviate the above security concerns.

\textbf{Feature security}:
The active party has the following information about the passive party: feature dimension $d$, sample space $I$, and accumulated label sum by percentile  $ \langle \mathbf{Y} \rangle$. 
Generally speaking, it is not sufficient for the active party to derive the feature information on such information. 
% {\color{red}However, under some extreme cases, there still exists a certain risk: Suppose there is only one active and passive party, active party has no feature information and passive party only has one feature.}
% The active party randomly generate a set of label to start training, and the passive party will send the sample space of the left and right subtree to the active party.
{ However, if the active party keeps sending a small sample space to the passive party, the feature rank in the passive party could be learned by the active party.}
A simple strategy that can be used to protect passive party's feature information is to randomly shuffle the left and right subtree while splitting. In this way, the active party will not be able to detect the passive party feature information even under the extreme case we constructed above.
%%%%%%%%%%%%%%%%%%%%%%%%%%%%%%%%%%%%%%%%%%%%%%%%%%%%%%%%%%%%%%%%%%%%%%%%%%%%%%%
%%%%%%%%%%%%%%%%%%%%%%%%%%%%%%%%%%%%%%%%%%%%%%%%%%%%%%%%%%%%%%%%%%%%%%%%%%%%%%%

\section{Cryptosystem}
We choose RIAC as the cryptosystem in our work mainly due to its efficiency. RIAC is built on the classical affine cipher, which uses a linear transformation and a mode operation to encrypt and decrypt the plaintext:
\begin{equation}
\begin{aligned}
    E(x) & = (ax) \mod n \\
    D(x) & = (a^{-1} x) \mod n
\end{aligned}
\end{equation}
where $a$ and $n$ are coprime integers, $a^{-1}$ is the multiplicative inverse of $a$ under$\mod n$. Affine cipher is fast to compute; however, it has a major flaw as its ciphertext preserves the frequency property of plaintext to some extent, making it vulnerable under certain attacks. To this end, RIAC made two improvements: (1) iteratively apply multiple rounds of affine cipher encryption, (2) use auxiliary random number $r$ and $s$ to randomize the frequency of ciphertext:
\begin{equation}
\begin{aligned}
    E_{i+1}(x) & = \big( a_{i} E_i(x) \big) \mod n_{i} \\
    D_{i+1}(x) & = \big( a_{i}^{-1} D_{i}(x) \big) \mod n_{i}
\end{aligned}
\end{equation}
where $i$ is the current round of RIAC computation, and:
\begin{equation}
    \begin{aligned}
        E_{0}(x) & = [(y \cdot s)\mod n_{0},(x + y\cdot h) \mod n_{0}]\\
        h & = (s \cdot r) \mod n_{0}\\
        D_{0}(x) & = (c_2 - r \cdot c_1) \mod n_{0}\\
    \end{aligned}
\end{equation}
Here $c_1, c_2$ are two components of the ciphertext obtained from the last decryption round.
{And $y$ is an extra randomly generated large integer to further randomize the computation, so that every time the encoding of the same number will lead to different ciphertext.}

\begin{table*}
\caption{Comparison on the time consumption of RIAC and Paillier protocols under Python implementation. It is assumed that the dataset has 1 million samples in total.}
\label{table:message_compress_ratio}
\vskip 0.15in
\begin{center}
\begin{small}
\begin{sc}
\begin{tabular}{lcccccr}
\toprule
%\multicolumn{2}{}{Sample rate\\ \& Method} & mssage\\size (kB) & transfer\\cost (s) & serialize\\cost (s) & speedup\\ratio\\
%Protocol & Paillier ($\mu$s) & RIAC ($\mu$s) & Speedup \\ 
& Sample rate \& Method & mssg size (kB) & transfer cost (s) & serialize cost (s) & speedup ratio & best?\\
\midrule
\multirow{3}{0pt}{\rotatebox[origin=c]{90}{10\%}} &V0 & 788.9 & 5.136 & 5.329 & 1.0$\times$  & \\
&V1 & 333.3 & 2.170 & 2.340 & 2.3$\times$ & \\
&V2 & 112.2 & 0.732 & 0.874 & 6.1$\times$ & $\surd$\\
\hline 
\multirow{3}{0pt}{\rotatebox[origin=c]{90}{50\%}} &V0 & 394.4 & 25.680 & 26.034 & 1.0$\times$ & \\
&V1 & 333.3 & 2.170 & 2.296 & 11.3$\times$ & $\surd$\\
&V2 & 112.5 & 3.662 & 3.934 & 6.6$\times$ & \\
\hline
\multirow{3}{0pt}{\rotatebox[origin=c]{90}{90\%}} &V0 & 7888 & 51.360 & 52.208 & 1.0$\times$ & \\
&V1 & 333.3 & 2.170 & 2.272 & 23.0$\times$ & $\surd$\\
&V2 & 2125 & 13.834 & 14.151 & 3.7$\times$ & \\
\bottomrule
\end{tabular}
\end{sc}
\end{small}
\end{center}
\vskip -0.1in
\end{table*}

\section{Ablation Study}
\label{appendix:ablation_study}

A benchmark to communication optimization is done with the FK dataset, where the results are visualized in Fig. \ref{fig:optimized_communication}. Without any optimization in communication, the training would take 210 min to finish. With data compression applied, the training time is reduced to 120 min. With both compression and async communication, the time cost is further reduced to 60 min.

\begin{figure}
    \centering
    \includegraphics[width=6cm]{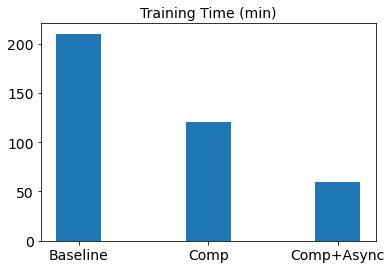}
    \caption{Benchmark on optimized communication. "Baseline", "Comp", and "Comp+Async" denotes model without any communication optimization, model with data compression along, and model with both data compression and asynchronous communication, respectively.}
    \label{fig:optimized_communication}
\end{figure}

\subsection{Optmized Communication}

A comparison of the speed difference for the above methods is listed in Tab. \ref{table:message_compress_ratio}. The result is obtained with different sampling rates on 1 million samples, and the network speed is assumed to be 300 kB/s. It can be seen that bit encoding performs the best under large and medium sampling rates, while extreme bit encoding performs the best under small sampling rates. And both methods are much faster than the vanilla approach under all settings.
In our final system, we choose to transfer the data index through bit encoding when the sampling rate is larger than 15\%, and through extreme bit encoding otherwise.

To understand the scalability with more participating parties, we conduct the following experiment on the Epsilon dataset: Assuming the features are evenly distributed on all parties, we increase the number of participating parties from 3 to 5 to 10 and train the system. Without the proposed O3 (optimized communication), the training time is 66.1 min, 129.4 min, 292.6 min, respectively. However, with O3, the training time becomes 23.7 min, 27.6 min, and 32.5 min, which is 2.79x, 4.69x, and 9.00x faster. This empirically shows that the proposed O3 is not only helpful at reducing the time consumption at the same number of parties but also significantly reduces the growth of time consumption to sublinear. Thus making the system more scalable in terms of participating parties.

\subsection{RIAC}

As a comparison, Tab. \ref{table:RIAC_python_result} shows the speedup of RIAC over the popular Paillier cryptosystem \cite{paillier1999public}. The time consumption is obtained from an average of 1000 plaintext numbers randomly drawn from \(N(0,10^8)\). As a reference, similar addition and multiplication operations in plaintext take 7.15 ns and 7.22 ns, respectively. It can be clearly seen that, although both RIAC and Paillier are much more time consuming than plaintext operations, RIAC is much faster than Paillier and can help to reduce the overall training time of federated random forest.

\begin{table}
\caption{Comparison on the time consumption of RIAC and Paillier protocols at performing different computations.}
% under Python implementation.}
\label{table:RIAC_python_result}
\vskip 0.15in
\begin{center}
\begin{small}
\begin{sc}
\begin{tabular}{lcccr}
\toprule
Protocol & Paillier ($\mu$s) & RIAC ($\mu$s) & Speedup \\ 
\midrule
scalar mult. & \(53.17\) & \(3.66\) & \(14 \times\)\\
addition & \(148.38\) & \(2.54\) & \(58 \times\)\\
decryption & \(4.76\) & \(0.0036\) & \(130 \times\)\\
encryption & \(16570\) & \(53.86\) & \(307 \times\)\\
\bottomrule
\end{tabular}
\end{sc}
\end{small}
\end{center}
\vskip -0.1in
\end{table}

\subsection{MPC}
As an example, we compute $(c_1\cdot c_2)\mod n$ and $(c_1^{10}) \mod n$, where $c_1,c_2,n$ are big integers of 300 digits. The speed comparison of mulmod and powmod operators with and without GMP is shown in Tab. \ref{table:gmp_performance}. Although the actual run time may vary with different integer digits and machines, it is obvious that GMP could make the cryptosystem much faster.

\begin{table}
\caption{GMP Benchmark on mulmod and powmod operations.}
\label{table:gmp_performance}
\vskip 0.15in
\begin{center}
\begin{small}
\begin{sc}
\begin{tabular}{lcccr}
\toprule
Op.  & w. GMP($\mu s$) & w/o GMP ($\mu s$)  & Speedup \\
\midrule
mulmod  & 1.17 & 4.67 & 4$\times$ \\
powmod & 3.95  & 60.8 & 15$\times$ \\
\bottomrule
\end{tabular}
\end{sc}
\end{small}
\end{center}
\vskip -0.1in
\end{table}

\end{document}